%% file: iclr2024_conference.tex
\documentclass{article} 
\usepackage{iclr2024_conference,times}

\input{math_commands.tex}

\usepackage{hyperref}
\usepackage{url}
\usepackage{graphicx}
\usepackage{longtable}
\usepackage[section]{placeins}
\usepackage{tabularx}
\usepackage{multirow}
\usepackage{verbatim}
\usepackage{float}
\usepackage{authblk}
\usepackage{amsmath}

\begin{document}
\title{Explaining latent representations of generative models with large multimodal models}


\author[1]{Mengdan Zhu}
\author[1]{Zhenke Liu}
\author[1]{Bo Pan}
\author[2]{Abhinav Angirekula}
\author[1]{Liang Zhao}
\affil[1]{Department of Computer Science, Emory University}
\affil[2]{Department of Computer Science, University of Illinois Urbana-Champaign}
\affil[ ]{\texttt {\{mengdan.zhu, zhenke.liu, bo.pan, liang.zhao\}@emory.edu},
\texttt{aa125@illinois.edu}}

\renewcommand\Authands{, }

%


\iclrfinalcopy 

\maketitle

\begin{abstract}
Learning interpretable representations of data generative latent factors is an important topic for the development of artificial intelligence. With the rise of the large multimodal model, it can align images with text to generate answers. In this work, we propose a framework to comprehensively explain each latent variable in the generative models using a large multimodal model. We further measure the uncertainty of our generated explanations, quantitatively evaluate the performance of explanation generation among multiple large multimodal models, and qualitatively visualize the variations of each latent variable to learn the disentanglement effects of different generative models on explanations. Finally, we discuss the explanatory capabilities and limitations of state-of-the-art large multimodal models.
\end{abstract}

\section{Introduction}

Latent variable based data generation has emerged as a state-of-the-art (SOTA) approach in the field of generative modeling \citep{pmlr-v202-mittal23a,deja2023learning,patil2022dot}. This technique leverages latent variables to learn underlying data distributions effectively and also generate high-quality samples \citep{vahdat2021score}. One of the key advantages of using latent variables is their ability to capture the underlying structure in high-dimensional data. However, understanding and interpreting such latent variables is challenging and often requires human expertise for meaningful interpretation. Learning interpretable representations of the data generative latent factors is an important topic for the development of artificial intelligence that is able to learn and reason the same as humans do \citep{higgins2016beta}. Large multimodal models (LMMs) have accomplished remarkable progress in recent years \citep{wang2024exploring,wu2023multimodal,yin2023survey,bai2024beyond,ling2023domain}. LMM is more similar to the way humans perceive the world \citep{yin2023survey}. We thus consider using large multimodal models to automatically explain the latent representations.


Recently, one of the most powerful LMMs is the instruction-following LMM \citep{li2023large}. LLaVA and InstructBLIP are two instruction-following LMMs that achieve SOTA performance on many datasets. Instruction-following models use instruction tuning to enhance their abilities to understand and follow human-given instructions. Instruction tuning involves further fine tuning LLMs using $\langle instruction, response \rangle$ pairs to better align human intent with model behavior \citep{wang2024exploring}. LLaVA was introduced in the paper Visual Instruction Tuning \citep{liu2023visual}, and then further improved in Improved Baselines with Visual Instruction Tuning (referred to as LLaVA-1.5) \citep{liu2023improvedllava}. Likewise, InstructBLIP \citep{dai2023instructblip} is a large multimodal model that adds instruction tuning on the basis of its previous version BLIP-2 \citep{li2023blip2}. Google Bard is a conversational AI service developed by Google, initially powered by LaMDA with a range of models to follow. In this work, we propose a framework to comprehensively explain each latent variable in generative models and evaluate the performance of explanation generation of GPT-4-vision with several popular LMMs: Google Bard, LLaVA-1.5, and InstructBLIP. To the best of our knowledge, we are the first to use LMM to explain the latent representations of the generative models.

\section{Methods}
\textbf{Problem Formulation}
We start from a dataset $\mathcal{D}=\{X, Z\}$ where the images $\mathbf{x} \in \mathbb{R}^N$ and the data generative latent variables $\mathbf{z} \in \mathbb{R}^M$. We then can train a generative model that learns the joint distribution of the data $\mathbf{x}$ and latent variables $\mathbf{z}$. Here, our goal is to explain each latent variable $\evz_{i}$ individually, where i ranges from 1 to the number of latent dimensions, and quantify the uncertainty of the explanation to make sure the explanations presented are reliable and responsible. 

In this work, we confront two primary challenges associated with the representation and interpretation of the latent space. Firstly, the latent space is difficult to explicitly represent. To address this, we interpolate a specific latent dimension $\evz_{i}$ at a time and subsequently decode these into an image sequence. This method allows us to visualize the latent representations. Secondly, not all latent variables have semantic meaning. To tackle this issue, we introduce an uncertainty approach to distinguish between interpretable and uninterpretable latent variables.


\begin{figure}[t]
\begin{center}
\includegraphics[width=\textwidth]{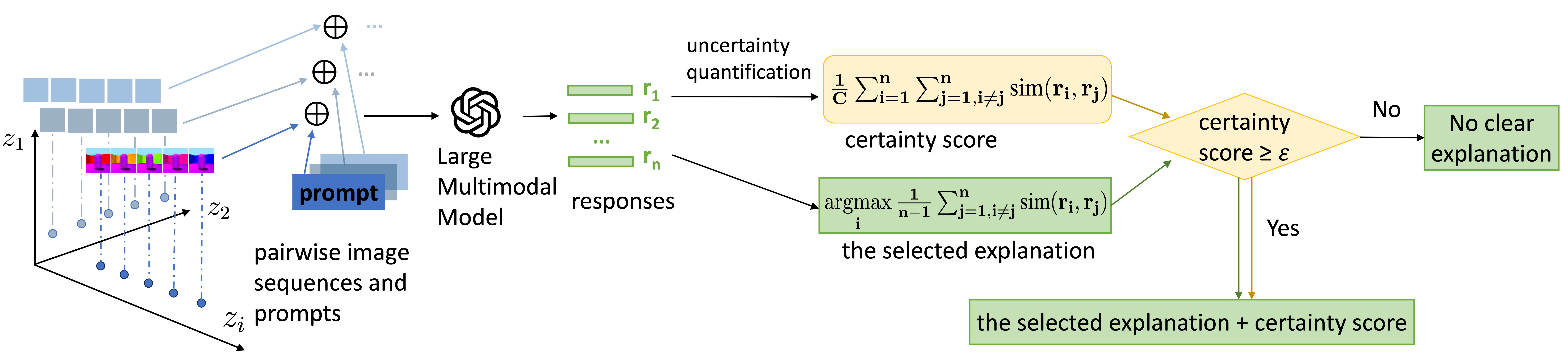}
\caption{The model framework consists of generating an image sequence with a progressively varying latent variable, combining it with a prompt to pass to a large multimodal model to provide some response samples and finally utilizing an uncertainty measure to select an explanation for that specific latent variable and decide whether there is a clear explanation to display.}
\label{overview}
\end{center}
\end{figure}

\textbf{Framework}
To start with, we trained three generative models for each dataset. After the generative models are trained, a latent representation $\rvz$ is first sampled from an isotropic unit Gaussian distribution $\mathcal{N}(\mathbf{0},\mI)$. Given a certain latent variable $\evz_{i}$, we perturb $\evz_{i}$ to observe the possible value range of $\evz_{i}$ and keep the other latent variables constant. A sequence of images can then be generated by decoding a series of manipulated latent vectors. As a result, this generated sequence of images shows how a latent variable $\evz_{i}$ changes gradually. Further, we pass this image sequence with a prompt to a large multimodal model to explain the latent variable, as well as its changing pattern. Finally, we quantify the uncertainty to decide which response to select as an explanation and whether to display the selected explanation as shown in Figure~\ref{overview}. 

Not every latent variable is interpretable, so we need to find an approach to determine whether a latent variable is interpretable. Similarity and entropy are two major ways to measure uncertainty in natural language generation(NLG). Since the likelihood of tokens is not available in GPT-4-vision, we used the measure of similarity in this paper. However, the use of similarity is not a limitation of the uncertainty measure. Other methods, such as predictive entropy, semantic entropy \citep{kuhn2023semantic}, and P(true) \citep{kadavath2022language}, could also be used to quantify uncertainty if applicable. To measure the uncertainty of the responses of our large multimodal model, GPT-4-vision, we sampled $n$ times from the GPT-4-vision to generate the responses $R=\{r_1,r_2,r_3,...,r_n\}.$ The certainty score of the explanation is the average similarity of the responses $R:\frac{1}{C}\sum_{i=1}^{n}\sum_{j=1, i \neq j}^n\text{sim}(r_i,r_j)$, where $C = n*(n-1)/2$. To find the threshold of the interpretability of the latent variables, we denote the true label of the interpretability of each latent variable as 1 if a human can see a clear pattern in the generated images, otherwise we denote it as 0. The certainty score of the explanation for each row of generated images is its predicted probability. We can then compute their AUC (area under the ROC curve) based on different thresholds and choose the one with the highest AUC as our threshold $\varepsilon$. Our final output explanation is the explanation that has the highest mean pairwise similarity with other responses if the certainty score is equal or greater than the threshold $\varepsilon$. Otherwise, we will output there is no clear explanation. 

\section{Experiments}
\textbf{Datasets}
We perform evaluations with three datasets, the MNIST dataset of handwritten digits \citep{lecun-mnisthandwrittendigit-2010}, dSprites dataset of 2D shapes \citep{dsprites17} and 3dshapes dataset of 3D shapes \citep{3dshapes18}. The dSprites dataset consists of 6 ground truth latent factors. These factors are color(white), shape(square, ellipse, heart), scale(6 values), rotation(40 values), position X(32 values) and position Y(32 values) of a sprite. Similarly, the 3dshapes dataset is generated from 6 ground truth latent factors of floor color(10 values), wall color(10 values), object color(10 values), scale(8 values), shape(4 values), and orientation(15 values). The MNIST dataset consists of grayscale handwritten digits(0 through 9).

\textbf{Visual pattern generation}
We train three representative variational autoencoder(VAE) models, the standard VAE \citep{kingma2014auto,rezende2014stochastic}, $\beta$-VAE \citep{higgins2016beta}, and $\beta$-TCVAE \citep{chen2018isolating}, with three aforementioned datasets. For each trained model, we manipulate one latent variable between [-3, 3] at a time while keeping others constant, and the trained decoder of the model can produce a series of images that reflect variations along that specific latent dimension. This process is repeated for each latent variable, resulting in $m \times k$ images, where $m$ is the number of latent variables and $k$ is the number of assigned values for each latent variable. We set 6 latent variables and assign the values as \textit{torch.range(-3, 3, 0.6)}, so $m = 6$ and $k = 11$ here.

\textbf{Explanation generation}
We use GPT-4-vision as our explanation generator and compare it with a wide range of other large multimodal models: Google Bard, LLaVA-v1.5-13b \citep{liu2023improvedllava}, and InstructBLIP \citep{dai2023instructblip}. For GPT-4-vision, LLaVA-v1.5-13b, and InstructBLIP, we all set $temperature = 1, top\_p = 1$. We then pass the generated images of each latent variable along with a prompt to the LLMs to produce an explanation for each latent variable. For human annotations, two annotators provide two explanations with various sentence expressions for each image sequence, so we have four human explanations for each image sequence as references. Furthermore, We evaluate the explanations across these LMMs with human annotations using BLEU\citep{papineni-etal-2002-bleu}, ROUGE-L\citep{lin-2004-rouge}, METEOR\citep{banerjee-lavie-2005-meteor}, and BERTScore F1\citep{zhang2019bertscore}. 

Additionally, to compute the certainty scores, we try both the cosine similarity and lexical similarity \citep{kuhn2023semantic}, where $sim$ is cosine similarity and rouge-L respectively in $\frac{1}{C}\sum_{i=1}^{n}\sum_{j=1, i \neq j}^n\text{sim}(r_i,r_j)$. As Table~\ref{rule} shows, the measure of cosine similarity has a much higher AUC, which means it can better distinguish if there is a clear pattern. So we choose to use the cosine similarity to generate the certainty score for the explanation. The best AUC in the experiment is 0.9694 and its corresponding threshold $\varepsilon$ is 0.7434.

\begin{table}[h]
\begin{center}
\caption{The evaluation metrics of the classification of the presence of a discernible trend.}
\label{rule}
\begin{tabularx}{\textwidth}
{ 
  | >{\centering\arraybackslash}X
  | >{\centering\arraybackslash}p{2cm} 
  | >{\centering\arraybackslash}p{2cm} 
  | >{\centering\arraybackslash}p{2cm} 
  | >{\centering\arraybackslash}p{2cm} | }
 \hline
 \textbf{Uncertainty Estimate} & \textbf{AUC} & \textbf{F1-score} & \textbf{Precision} & \textbf{Recall}\\ 
 \hline
  lexical similarity & 0.6898  & 0.9600  & 0.9412 & \textbf{0.9796}\\
  \hline
 cosine similarity & \textbf{0.9694}  & \textbf{0.9684}  & \textbf{1.0} & 0.9388 \\
\hline
\end{tabularx}
\end{center}
\end{table}

\section{Results and discussion} 
\textbf{Quantitative evaluation for explanation generation}
We give the same prompt and image sequence to GPT-4-vision and other large multimodal models to generate explanations. Overall, GPT-4-vision outperforms other large multimodal models in their explanatory capability to explain the visual patterns of latent variables in Table~\ref{evaluation-automated}. As shown in Appendix~\ref{example:mnist}, GPT-4-vision is the only LMM that can accurately recognize the handwritten digits in MNIST, explain how the digits change, and have the best overall explanations. Bard can answer the latent variable based on the prompt, but the latent variables and the patterns found are not always accurate. LLaVA and InstructBLIP are not able to respond based on the prompt. More specifically, LLaVA can only describe the images, but it cannot explicitly answer what the latent variable is. InstructBLIP often repeats the task content in the prompt, yet fails to provide the required explanations.

\begin{table}[t]
\caption{Evaluation of large multimodal models on the generated explanations of the latent variables of different generative models}
\label{evaluation-automated} 
\begin{tabular}{cclcccc}
\hline
\textbf{Dataset}           & \textbf{VAE Model}         & \textbf{LMM} & \textbf{BLEU} & \textbf{ROUGE-L} & \textbf{METEOR} & \textbf{BertScore} \\ \hline
\multirow{12}{*}{3dshapes} & \multirow{4}{*}{VAE}       & \textit{GPT-4-vision} & \textbf{0.051}         & \textbf{0.196 }           & \textbf{0.370}           & \textbf{0.875}              \\
                           &                            & \textit{Bard}         & 0.047         & 0.167            & 0.240           & 0.842              \\
                           &                            & \textit{LLaVA-1.5}    & 0.000         & 0.169            & 0.224           & 0.864              \\
                           &                            & \textit{InstructBLIP} & 0.000         & 0.167            & 0.212           & 0.843              \\ \cline{2-7} 
                           & \multirow{4}{*}{$\beta$-VAE}   & \textit{GPT-4-vision} & \textbf{0.056}         & \textbf{0.195 }           & \textbf{0.302}           & \textbf{0.868}              \\
                           &                            & \textit{Bard}         & 0.000         & 0.161            & 0.224           & 0.850              \\
                           &                            & \textit{LLaVA-1.5}    & 0.024         & 0.162            & 0.206           & 0.857              \\
                           &                            & \textit{InstructBLIP} & 0.027         & 0.160            & 0.205           & 0.842              \\ \cline{2-7} 
                           & \multirow{4}{*}{$\beta$-TCVAE} & \textit{GPT-4-vision} & 0.058         & \textbf{0.203 }           & \textbf{0.293}           & \textbf{0.865}              \\
                           &                            & \textit{Bard}         & 0.000         & 0.189            & 0.211           & 0.864              \\
                           &                            & \textit{LLaVA-1.5}    & 0.030         & 0.181            & 0.206           & 0.856              \\
                           &                            & \textit{InstructBLIP} & \textbf{0.062 }        & 0.133            & 0.107           & 0.846              \\ \hline
\multirow{12}{*}{dsprites} & \multirow{4}{*}{VAE}       & \textit{GPT-4-vision} & 0.051         & 0.190            & \textbf{0.291}           & 0.858              \\
                           &                            & \textit{Bard}         & 0.052         & \textbf{0.225}            & 0.197           & \textbf{0.864}              \\
                           &                            & \textit{LLaVA-1.5}    & 0.062         & 0.176            & 0.219           & 0.855              \\
                           &                            & \textit{InstructBLIP} & \textbf{0.065}         & 0.195            & 0.205           & 0.847              \\
                           \cline{2-7} 
                           & \multirow{4}{*}{$\beta$-VAE}   & \textit{GPT-4-vision} & \textbf{0.061}         & \textbf{0.222}            & \textbf{0.282}           & \textbf{0.867}              \\
                           &                            & \textit{Bard}         & 0.049         & 0.210            & 0.233           & 0.856              \\
                           &                            & \textit{LLaVA-1.5}    & 0.048         & 0.197            & 0.194           & 0.856              \\
                           &                            & \textit{InstructBLIP} & 0.042         & 0.190            & 0.225           & 0.842              \\
                           \cline{2-7} 
                           & \multirow{4}{*}{$\beta$-TCVAE} & \textit{GPT-4-vision} & 0.051         & 0.193            & \textbf{0.276}           & \textbf{0.858}              \\
                           &                            & \textit{Bard}         & 0.028         & \textbf{0.209}            & 0.262           & 0.848              \\
                           &                            & \textit{LLaVA-1.5}    & \textbf{0.052}         & 0.184            & 0.187           & 0.857              \\
                           &                            & \textit{InstructBLIP} & 0.041         & 0.194            & 0.249           & 0.849              \\ \hline
\multirow{12}{*}{MNIST}    & \multirow{4}{*}{VAE}       & \textit{GPT-4-vision} & \textbf{0.038}         & 0.181            & \textbf{0.291}           & 0.862              \\
                           &                            & \textit{Bard}         & 0.000         & \textbf{0.203}            & 0.237           & \textbf{0.865}              \\
                           &                            & \textit{LLaVA-1.5}    & 0.027         & 0.162            & 0.265           & 0.857              \\
                           &                            & \textit{InstructBLIP} & 0.000         & 0.144            & 0.217           & 0.843              \\ \cline{2-7} 
                           & \multirow{4}{*}{$\beta$-VAE}   & \textit{GPT-4-vision} & 0.000         & \textbf{0.201}            & \textbf{0.279}           & \textbf{0.863}              \\
                           &                            & \textit{Bard}         & \textbf{0.026}         & 0.184            & 0.208           & 0.857              \\
                           &                            & \textit{LLaVA-1.5}    & 0.000         & \textbf{0.201}            & 0.215           & 0.846              \\
                           &                            & \textit{InstructBLIP} & 0.000         & 0.156            & 0.209           & 0.841              \\ \cline{2-7} 
                           & \multirow{4}{*}{$\beta$-TCVAE} & \textit{GPT-4-vision} & 0.040         & 0.207            & \textbf{0.288}           & 0.863              \\
                           &                            & \textit{Bard}         & 0.041         & 0.227            & 0.240           & 0.857              \\
                           &                            & \textit{LLaVA-1.5}    & \textbf{0.099}         & \textbf{0.245 }           & 0.287           & \textbf{0.864}              \\
                           &                            & \textit{InstructBLIP} & 0.000         & 0.197            & 0.237           & 0.836              \\ \hline
\end{tabular}
\end{table}

\textbf{Uncertainty analysis}
We observe that when there is an evident pattern in the images, the certainty score of the corresponding explanation is likely to be high in Appendix \ref{clear_unclear_patterns} Figure \ref{clear_pattern}, showing the sampled explanations are more consistent. Conversely, when human can not find a clear pattern in the images, the certainty score is relatively low in Appendix \ref{clear_unclear_patterns} Figure \ref{unclear_pattern}, indicating the sampled explanations are more diverse. 

We quantitatively evaluate if the certainty threshold we find can correctly distinguish whether there is a clear trend in the generated image sequences in Table~\ref{rule}. Table~\ref{rule} demonstrates an overall excellent performance of the certainty threshold of cosine similarity to identify the cutoff for the presence of a discernible trend. In comparison, the uncertainty estimate of cosine similarity can better distinguish the scenarios when there is no clear trend than the uncertainty estimate of lexical similarity.

\textbf{Qualitative evaluation for disentanglement}
When latent factors are entangled in the image sequences, like the ones of VAE and $\beta$-VAE in Appendix \ref{fig:disentanglement}, the LMM typically can only identify one of the entangled latent factors in the explanations. Moreover, the greater entanglement of latent factors makes it more challenging to identify changes in all latent factors within an explanation and contributes to an increase in the uncertainty of the explanation, such as the latent factor of $\beta$-TCVAE is more disentangled and thus has a higher certainty score than the ones of VAE and $\beta$-VAE.  

\section{Case study} 
\begin{figure}[h]
\begin{center}
\includegraphics[width=\textwidth]{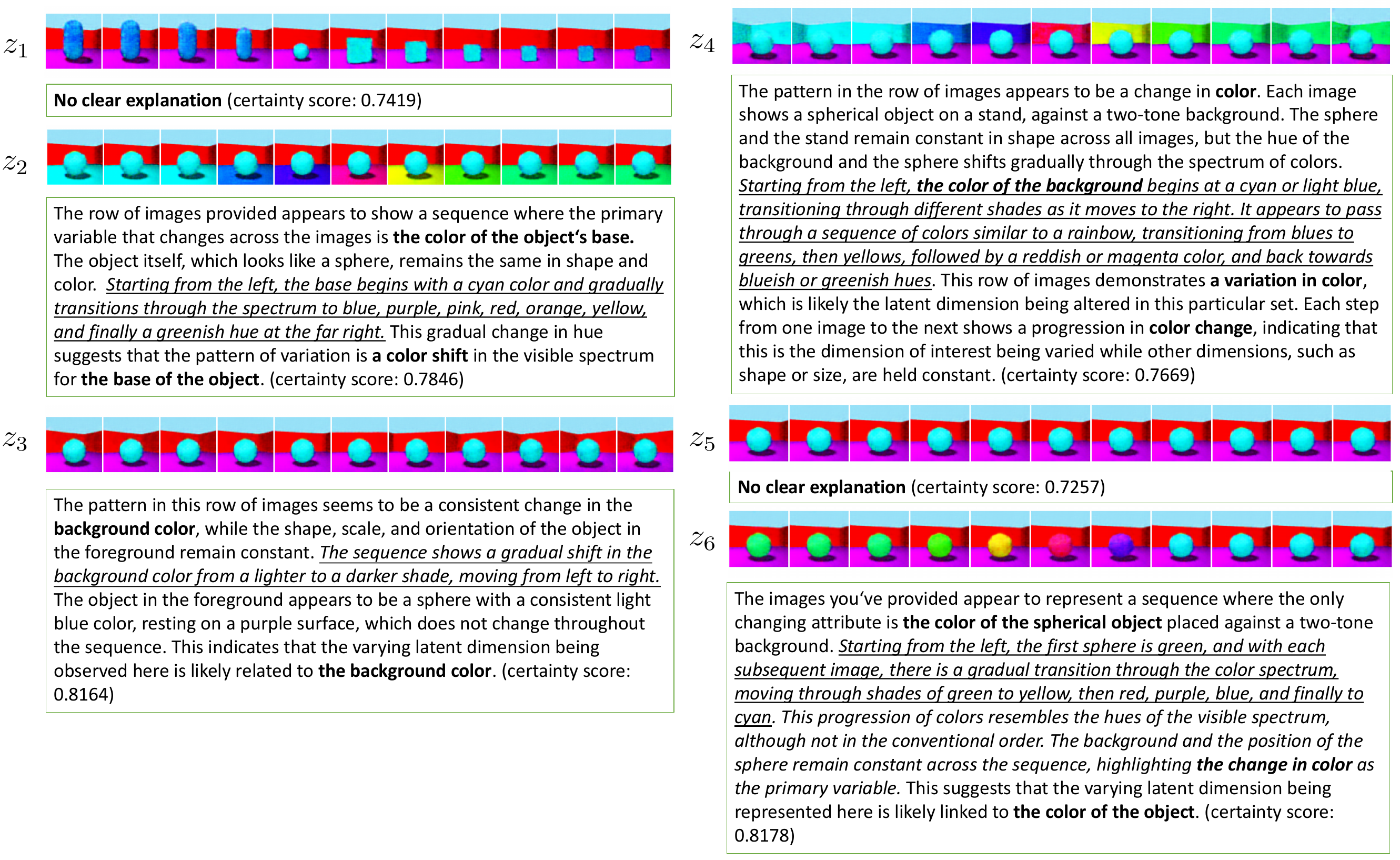}
\caption{The sample explanations generated by our framework. The latent variables are highlighted in \textbf{bold}, and the patterns of the latent variables are in \underline{\textit{italics and underlined}}.}
\label{fig:explanation_overview}
\end{center}
\end{figure}

Figure~\ref{fig:explanation_overview} shows the explanations generated by our framework for the 3dshapes dataset. If the certainty score of the generated explanation is equal or greater than the certainty threshold ($\varepsilon=0.7434$), it will display the explanation selected. Otherwise, it will return ``No clear explanation'' like $\evz_{1}$ and $\evz_{5}$. Our framework can generate an explanation for each latent variable if the latent variable has an evident pattern. The explanation not only can tell what the latent variable is, but also can illustrate how the latent variable changes. Sometimes, the identified latent variable can be wrong, such as it misclassifies the wall orientation as background color in $\evz_{3}$. This is due to the deficiency in the LMM's visual capabilities and will be elaborated in the later limitation section.

\section{Limitation}
Although GPT-4-vision is a state-of-the-art large multimodal model and performs the best in the experiments, we still find some of its limitations when evaluating its generated explanations. It is more likely to misinterpret the latent variable as color and is not sensitive to the scale, position, and orientation. Sometimes the description of color patterns is not entirely accurate. Therefore, more work can be undertaken to improve the visual understanding of GPT-4-vision.

\section{Conclusion}
In this work, we propose a framework to comprehensively explain each latent variable in the generative models and visualize the variations of each latent variable. We first analyze the certainty scores of the explanations. The certainty score can tell us if there is a clear trend in the latent variable. We also evaluate and compare the generated explanations with various LMMs. The result shows that GPT-4-vision outperforms other large multimodal models, and we further discuss its limitations and the explanatory capabilities of various LMMs. At last, we evaluate the effects of latent variable disentanglement on the generated explanations. We believe our approach provides an efficient, explainable, and reliable way to learn the latent representations of generative models.

\section*{Acknowledgments}
This work is supported by the National Science Foundation (NSF) Grant No. 1755850, No. 1841520, No.
2007716, No. 2007976, No. 1942594, No. 1907805, a
Jeffress Memorial Trust Award, Amazon Research Award,
Oracle for Research Grant Award, Cisco Faculty Research
Award, NVIDIA GPU Grant, Design Knowledge Company (subcontract number: 10827.002.120.04), CIFellowship (2021CIF-Emory-05), and the Department of Homeland Security under Grant No. 17STCIN00001.

\bibliography{iclr2024_conference}
\bibliographystyle{iclr2024_conference}

\newpage
\appendix
\section{Examples of explanation generation with clear and unclear patterns}
\label{clear_unclear_patterns}
\begin{figure}[h]
\begin{center}
\includegraphics[width=\textwidth]{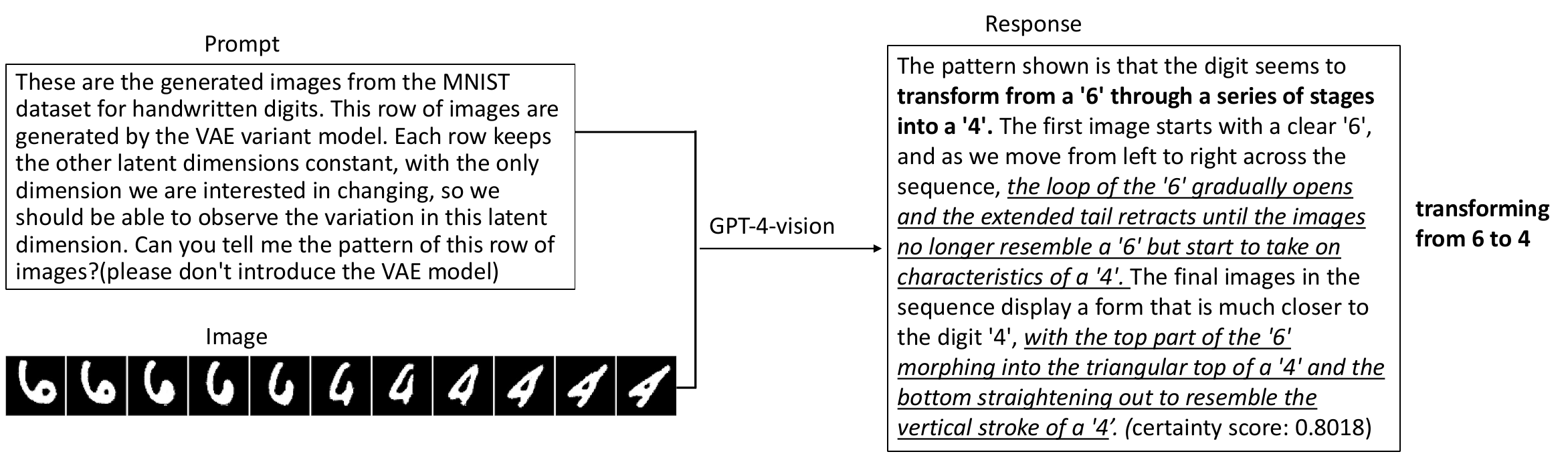}
\includegraphics[width=\textwidth]{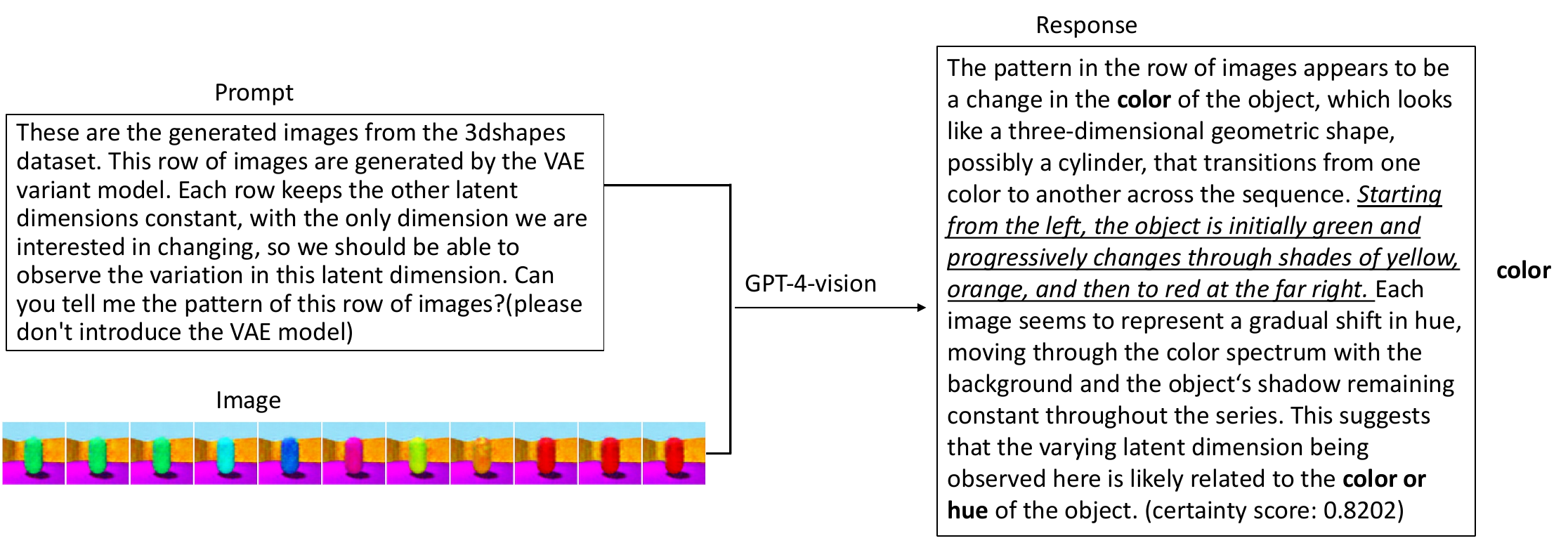}
\includegraphics[width=\textwidth]{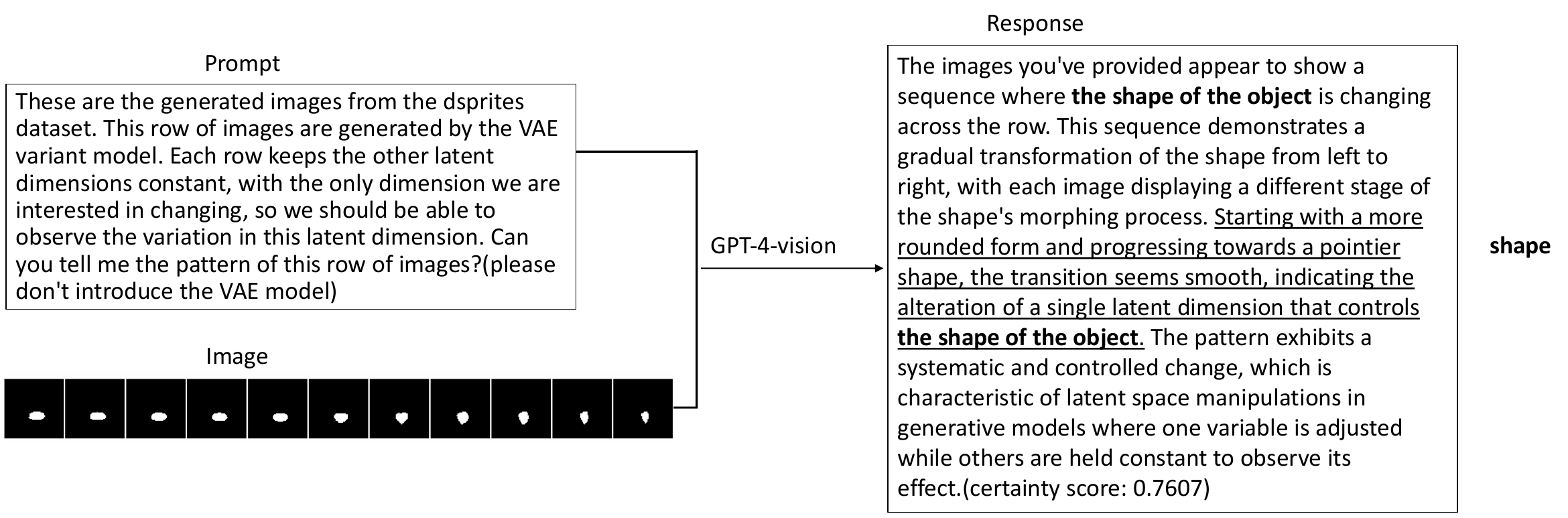}
\caption{Sample images with clear patterns and sample prompts for GPT-4-vision to generate explanations}
\label{clear_pattern}
\end{center}
\end{figure}

\begin{figure}[h]
\begin{center}
\includegraphics[width=\textwidth]{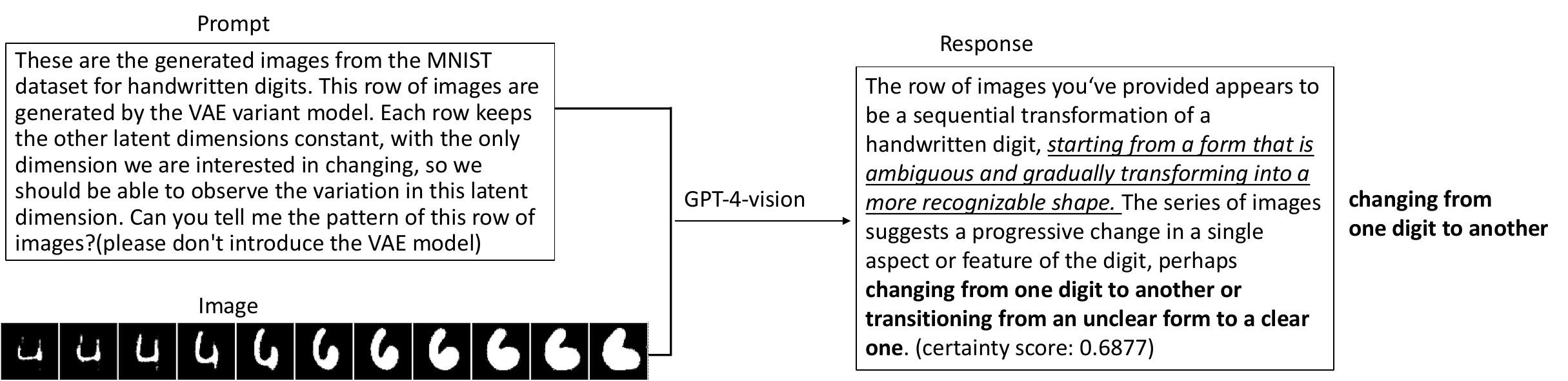}
\includegraphics[width=\textwidth]{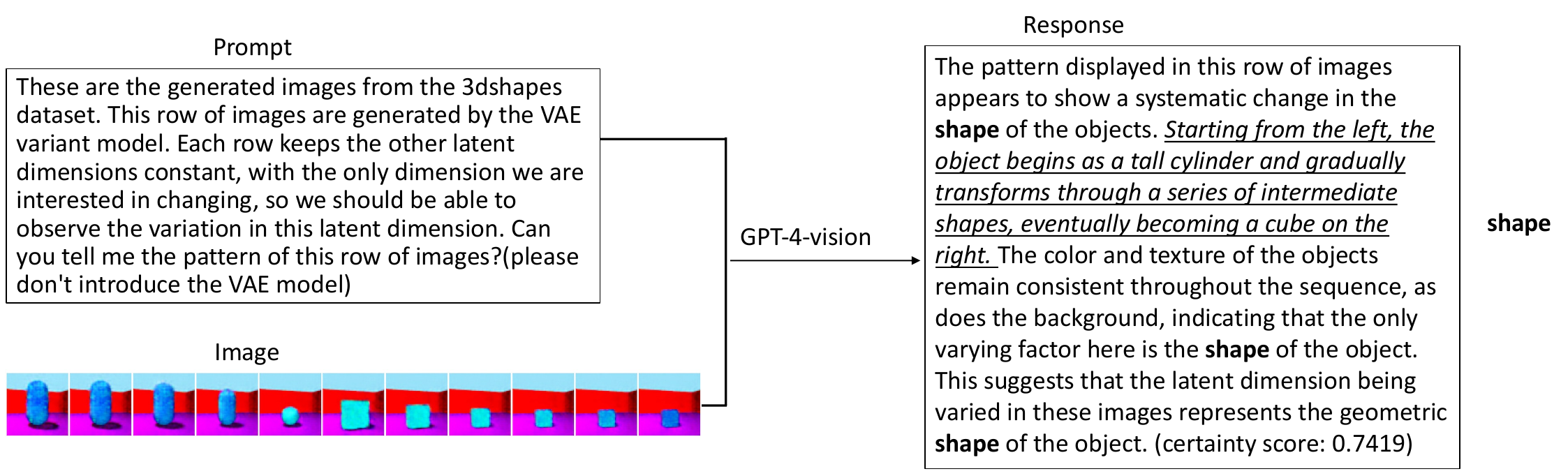}
\caption{Sample images with unclear patterns and sample prompts for GPT-4-vision to generate explanations}
\label{unclear_pattern}
\end{center}
\end{figure}

\section{Examples of explanations generated by different LMMs}
\label{example:mnist}
\begin{figure}[h]
\begin{center}
\includegraphics[width=\textwidth]{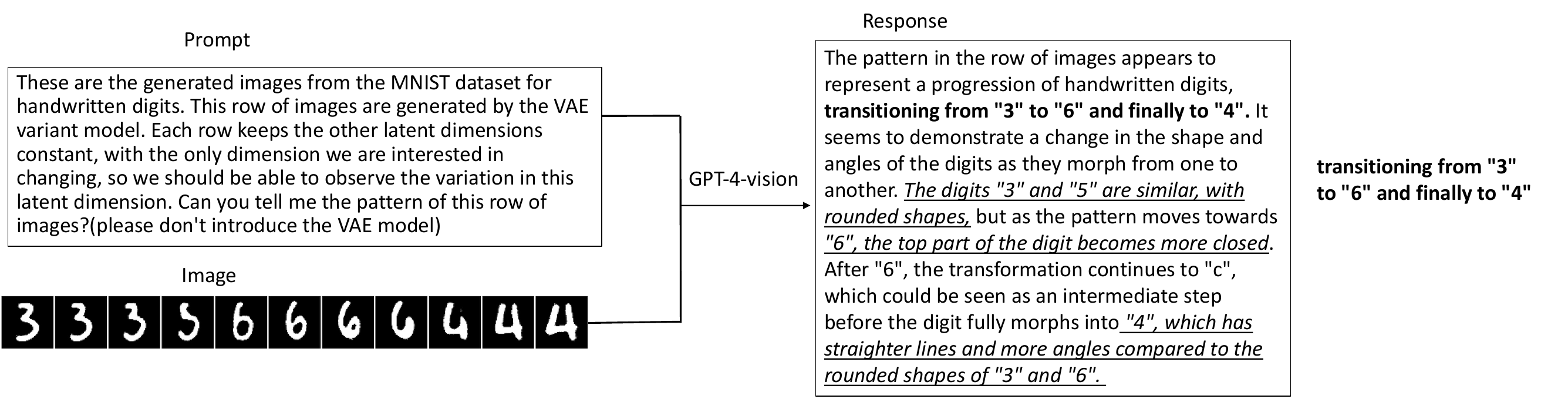}
\end{center}
\caption{The sample explanation generated by GPT-4-vision for the MNIST dataset.}
\end{figure}

\begin{figure}[h]
\begin{center}
\includegraphics[width=\textwidth]{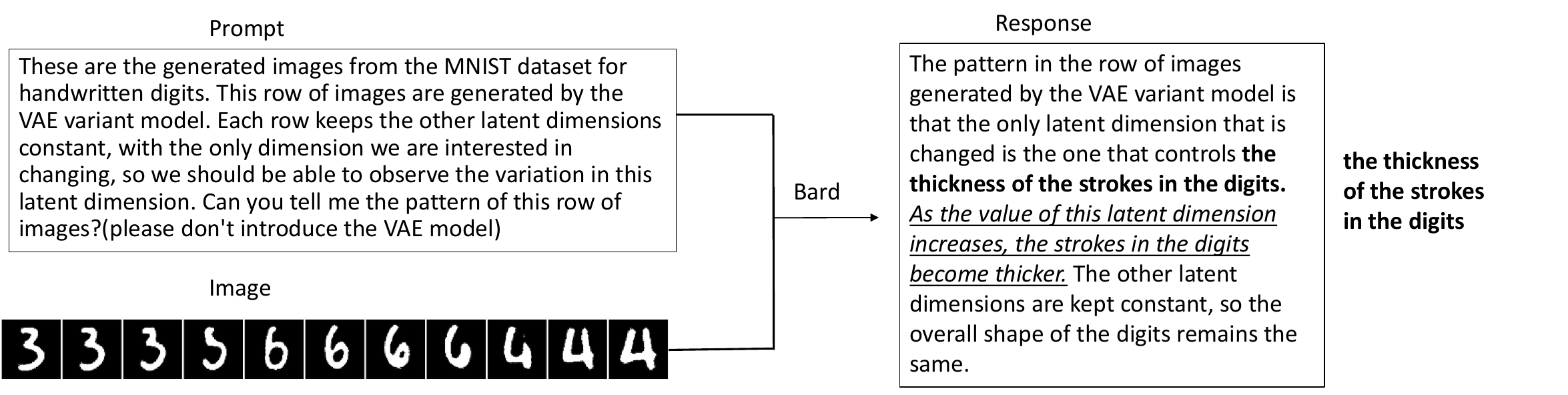}
\end{center}
\caption{The sample explanation generated by Bard for the MNIST dataset.}
\end{figure}

\begin{figure}[h]
\begin{center}
\includegraphics[width=\textwidth]{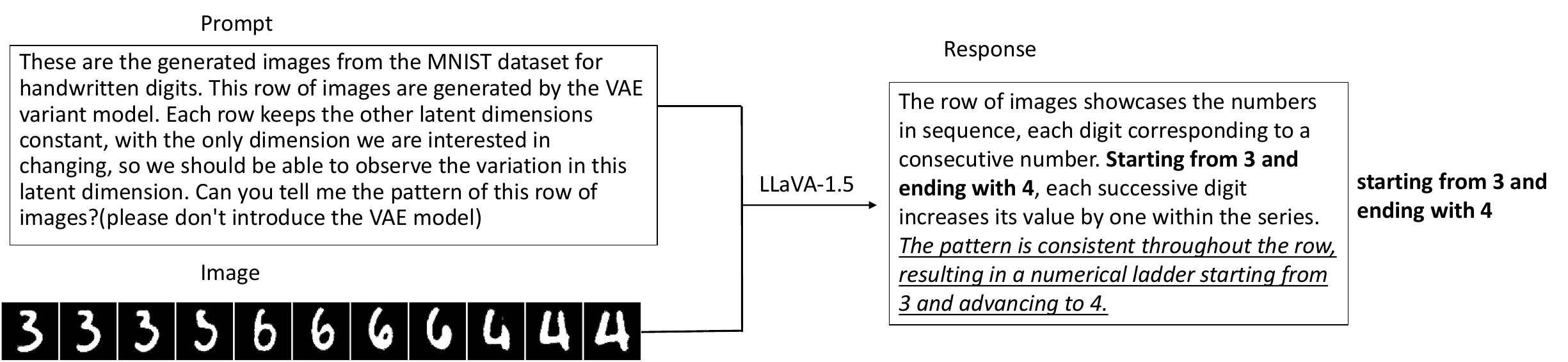}
\end{center}
\caption{The sample explanation generated by LLaVA-1.5 for the MNIST dataset.}
\end{figure}

\begin{figure}[h]
\begin{center}
\includegraphics[width=\textwidth]{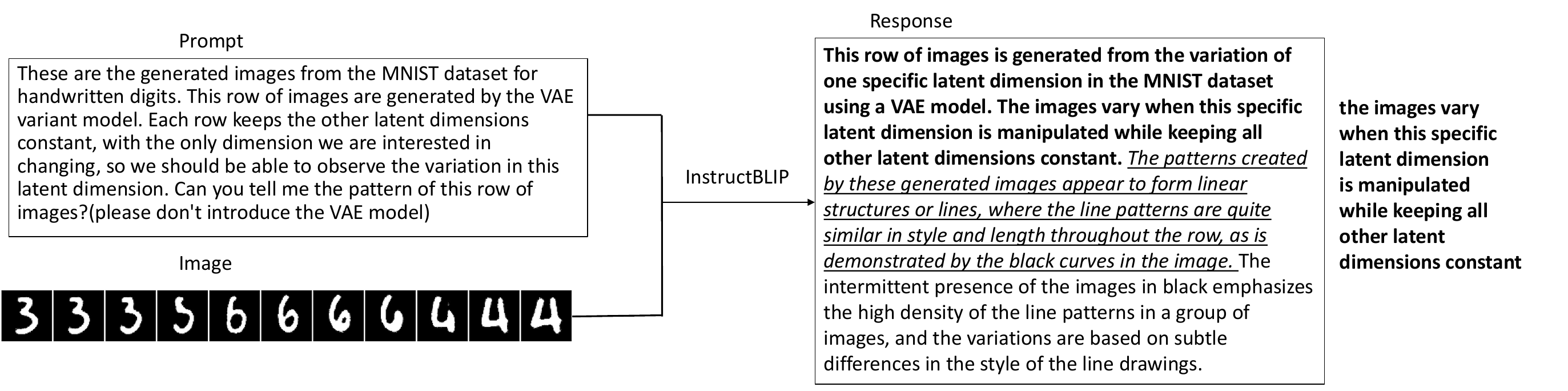}
\end{center}
\caption{The sample explanation generated by InstructBLIP for the MNIST dataset.}
\end{figure}

\section{The disentanglement effects of different generative models on explanations}
\label{fig:disentanglement}
\begin{figure}[H]
\begin{center}
\includegraphics[width=\textwidth]{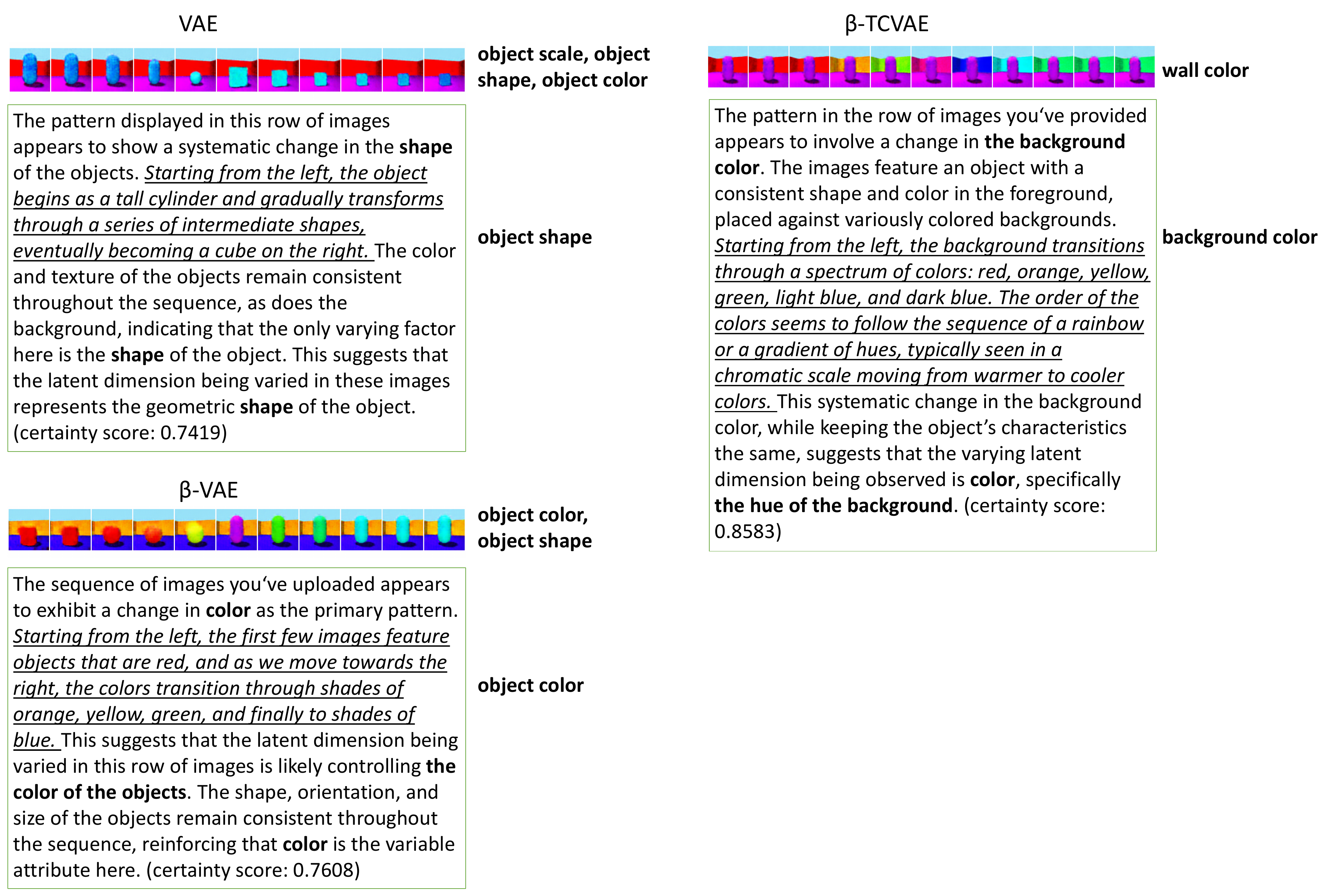}
\caption{The impact of disentanglement in different generative models on explanations. The ground truth latent factors in the image sequences and the latent variables in the explanations are highlighted in \textbf{bold}, and the patterns of the latent variables are in \underline{\textit{italics and underlined}}.}
\label{fig:disentanglement_discussion}
\end{center}
\end{figure}

\end{document}

%% file: math_commands.tex
\usepackage{amsmath,amsfonts,bm}

\def\eqref#1{equation~\ref{#1}}

\def\1{\bm{1}}

\def\rvz{{\mathbf{z}}}

\def\evz{{z}}

\def\mI{{\bm{I}}}

\DeclareMathAlphabet{\mathsfit}{\encodingdefault}{\sfdefault}{m}{sl}
\SetMathAlphabet{\mathsfit}{bold}{\encodingdefault}{\sfdefault}{bx}{n}

